\documentclass[11pt]{article}
\usepackage{coling2018}
 \usepackage{slashbox}
\usepackage{times}
\usepackage{latexsym}
 \usepackage{multirow}
\usepackage{amsmath,bm}
\usepackage{hhline}
\usepackage{times} %Required
\usepackage{helvet} %Required
\usepackage{courier} %Required
\usepackage{url} %Required
\usepackage{graphicx} %Required
\usepackage{times}
\usepackage{latexsym}
\usepackage{subcaption}
 
 \usepackage{amsmath}
\usepackage{amssymb}
\usepackage{tikz}
\usepackage{caption}
\usepackage{tabularx}
\usepackage{bm}
 \usepackage[utf8]{inputenc}
\usepackage{enumitem}
 
\usepackage{url}

 \title{On Adversarial Examples for Character-Level Neural Machine Translation}
\author
       {Javid Ebrahimi, Daniel Lowd, Dejing Dou 
       \\
       Computer and Information Science Department, University of Oregon, USA\\
       \{\texttt{javid, lowd, dou}\}@cs.uoregon.edu\\ 
       }
\begin{document}
\maketitle
\begin{abstract}

%NLP systems are increasingly being deployed and appreciated with relatively little attention to their vulnerabilities. We propose an efficient method to generate adversarial examples that would trick a deep character-aware model. We ask the question, ``How easy is it for an NLP system to break?'' Our experiments demonstrate that deep character-based models for text classification and machine translation are very brittle, and would behave undesirably after just a few character manipulations. We then ask the question, ``How expensive is it to break a model?'' We devise a novel fast gradient-based optimization method, which, given a budget of $r$ manipulations, requires an $\mathcal{O}(r)$ number of queries to the model.

 Evaluating on adversarial examples has become a standard procedure to measure robustness of deep learning models. Due to the difficulty of creating white-box adversarial examples for discrete text input, most analyses of the robustness of NLP models have been done through black-box adversarial examples. 
 We investigate adversarial examples for character-level neural machine translation (NMT), and contrast black-box adversaries with a novel white-box adversary, which employs differentiable string-edit operations to rank adversarial changes. We propose two novel types of attacks which aim to remove or change a word in a translation, rather than simply break the NMT. We
 demonstrate that white-box adversarial examples are significantly stronger than their black-box counterparts in different attack scenarios, which show more serious vulnerabilities than previously known. 
In addition, after performing adversarial training, which takes only 3 times longer than regular training, we can improve the model's robustness significantly.
%We analyze the properties of these examples, and show that employing these adversarial examples in training can improve test-time accuracy on clean examples, as well as defend the models against adversarial examples. 

%We further analyze the properties of these examples and whether or not examples designed to deceive one model are effective against another.

%Adversarial examples are synthetic examples, which are often constructed by manipulating real-world examples in order to make a classifier believe they belong to an incorrect class with high confidence \cite{goodfellow2014explaining}.
%Research on adversarial examples for neural nets has largely focused on image data, with relatively little work on textual data.
%Research on adversarial examples for neural nets has largely focused on image data, with relatively little work on textual data.

\end{abstract}

\section{Introduction}
\blfootnote{
    %
    % for review submission
    %
    \hspace{-0.65cm}  % space normally used by the marker
   This work is licensed under a Creative Commons Attribution 4.0 International Licence. Licence
details: http://creativecommons.org/licenses/by/4.0/.
    %
    % % final paper: en-uk version 
    %
    % \hspace{-0.65cm}  % space normally used by the marker
    % This work is licensed under a Creative Commons 
    % Attribution 4.0 International Licence.
    % Licence details:
    % \url{http://creativecommons.org/licenses/by/4.0/}.
    % 
    % % final paper: en-us version 
    %
    % \hspace{-0.65cm}  % space normally used by the marker
    % This work is licensed under a Creative Commons 
    % Attribution 4.0 International License.
    % License details:
    % \url{http://creativecommons.org/licenses/by/4.0/}.
}

Last year, a mistranslation by Facebook's machine translation (MT) system led to a wrongful arrest~\cite{haaretz17}. Instead of translating an Arabic phrase to ``good morning,'' Facebook's MT translated it as ``attack them.'' Arabic is a morphologically-rich language, and the MT mistook the input word for another which differs from the input by only one character. As MT is used more and more, it is increasingly important to understand its worst-case failures to prevent incidents like this.

Adversarial examples are inputs designed to make a machine learning model perform poorly, and are often constructed by manipulating real-world examples~\cite{goodfellow2014explaining}. Belinkov and Bisk~\shortcite{belinkov2017synthetic} investigate the sensitivity of neural machine translation (NMT) to synthetic and natural noise containing common misspellings. They show that state-of-the-art models are vulnerable to adversarial attacks even after a spell-checker is deployed. By performing ensemble adversarial training \cite{tramer2017ensemble}, they improve an NMT's robustness to adversarial noise.

We explore the space of adversarial examples for NMT in two directions: first, we study untargeted adversarial examples in a \textit{white-box} setting, wherein the adversary has access to model parameters and can use its gradients to inflict more damaging manipulations for a larger decrease in the BLEU score; second, equipped with the developed machinery to do white-box attacks, we can perform more interesting attacks.
%While an adversarial example is required to be similar to its corresponding clean example, real-world adversary for machine translation might have the additional requirement that the new translation should {not} be drastically different from the original translation. 
We propose \textit{controlled} and \textit{targeted} adversaries which create adversarial examples with other goals, instead of merely decreasing the BLEU score. A controlled adversary aims to mute a word in the original translation, while a targeted adversary aims to push a word into it. Table \ref{exmp-trans-targ} shows one example of each category. In both cases, the adversary, which has no word alignment model, has not drastically changed the rest of the translation, and has been able to reach its goals.

%Adversarial examples serve several purposes. First, they expose regions of the input space where the model performs poorly, which can aid in understanding and improving the model. By using these examples as training data, adversarial training learns models that are more robust and often perform better on non-adversarial examples. Furthermore, adversarial examples demonstrate vulnerabilities that could be exploited by malicious adversaries in real applications. 

  %In other settings, users of a social network may want to disguise their interests or demographics, in order to confuse biased ad placement algorithms, government surveillance, or even censorship. 
  
%Research on adversarial examples for neural nets has largely focused on image data; in the past year however, 
There is growing interest in understanding vulnerabilities of NLP systems \cite{jia2017adversarial,zhao2017generating,belinkov2017synthetic}. Previous work in NLP has focused on creating adversarial examples in a \textit{black-box} setting, wherein the attacker can query a model but does not have access to its parameters. 
%We investigate how an attacker can change the output of a system in a \textit{white-box} setting, wherein the adversary has access to model parameters. 
%The central question we answer is whether we can perturb discrete textual data to trick a model, while remaining comprehensible to humans. 
Black-box attacks often rely on heuristic methods to create adversarial examples. 
%Training models to be robust to these attacks does not guarantee robustness to any other attacks. 
In contrast, white-box attacks approximate the \emph{worst-case} attack for a particular model and input, within some allowed set of perturbations.  Therefore, white-box attacks can demonstrate and defend against a model's most serious vulnerabilities, which may not be discovered by black-box heuristics.
After exploring the space of adversarial examples for NMT and proposing new types of attacks, we focus on adversarial training to make our model more robust. 
 We build on HotFlip \cite{ebrahimi2017hotflip}, a recently-introduced white-box method for generating adversarial examples and performing adversarial training for text classification.
%Our method to generate adversarial examples is efficient, which makes it suitable to incorporate into adversarial training.
At the core of HotFlip lies an atomic {\em flip} operation, which changes one character to another by using the gradients of the model with respect to the one-hot vector input. In this work, we extend it to include a broader set of attacks, and we also improve it with a better beam search heuristic and faster adversarial training.   

\begin{table*}
% \vspace{-20pt}
\centering
{\renewcommand{\arraystretch}{0.8}
\scalebox{1}{
\begin{tabular}{p{1.4cm}p{14cm}}
\hline
 
 \small{src} & \footnotesize{ 1901 wurde eine Frau namens Auguste in eine medizinische Anstalt in Frankfurt gebracht.}\\
\small{adv} & \footnotesize{1901 wurde eine Frau namens \textbf{Afuiguste} in eine medizinische Anstalt in Frankfurt gebracht.}\\
\small{src-output} & \footnotesize{In 1931, a woman named \textbf{Augustine} was brought into a medical institution in France.}\\
\small{adv-output} & \footnotesize{In 1931, a woman named Rutgers was brought into a medical institution in France.}\\
% \small{src} & \footnotesize{ Wir erwarten Perfektion von Feministinnen, weil wir immer noch für so viel k{\"a}mpfen, so viel wollen, so verdammt viel brauchen.}\\
%\small{adv} & \footnotesize{Wir erwarten Perfektion von Feministinnen, weil wir immer noch für so viel k{\"a}mpfen, so viel wollen, so \textbf{öberdammt} viel brauchen.}\\
%\small{src-output} & \footnotesize{We expect perfection from feminist, because we're still fighting so much, so \textbf{damn}, so \textbf{damn}, so \textbf{damn}.}\\
%\small{adv-output} & \footnotesize{We expect perfection from feminist, because we still fight for so much, so much of all, so much of all that needs to be.}\\
 \hline
 
\small{src} & \footnotesize{Das ist Dr. Bob Childs -- er ist Geigenbauer und Psychotherapeut.}\\
\small{adv} & \footnotesize{Das ist Dr. Bob Childs -- er ist Geigenbauer und \textbf{Psy6hothearpeiut}.}\\
\small{src-output} & \footnotesize{This is Dr. Bob Childs -- he's a wizard maker and a \textbf{therapist}'s  \textbf{therapist}.}\\
\small{adv-output} & \footnotesize{This is Dr. Bob Childs -- he's a brick maker and a \textbf{psychopath}.}\\
\hline
\end{tabular}
}
}
%\vspace{-20pt}
\captionsetup{font=footnotesize}
\caption{Controlled and Targeted Attack on  DE$\rightarrow$EN NMT. In the first example, the adversary wants to suppress a person's name, and in the second example, to replace occurrences of \textit{therapist} with \textit{psychopath}}
\label{exmp-trans-targ}
\end{table*}

Our contributions are as follows:
\begin{enumerate}
%\item We propose an efficient gradient-based optimization method to manipulate discrete text structure at its one-hot representation, and create white-box adversarial examples.
\item We use a gradient-based estimate, which ranks adversarial manipulations, and we search for adversarial examples using greedy search or beam search methods.
\item We propose two translation-specific types of attacks and provide a metric to evaluate adversaries in these scenarios. Our experiments show that white-box adversaries can be significantly stronger than black-box adversarial examples.% We also study the human-discernibility of these adversarial examples.
\item We investigate the robustness of models trained with white-box adversarial examples and compare their robustness with black-box trained models.
%\item We study the transferability of these adversarial examples to another character-aware model for classification.
\end{enumerate}

\section{Related Work}
The need to understand vulnerabilities of NLP systems is only growing. Companies such as Google are using text classifiers to detect abusive language\footnote{https://www.perspectiveapi.com}, and concerns are increasing over deception \cite{zubiaga2016analysing} and safety \cite{chancellor2016thyghgapp} in social media. In all of these cases, we need to to better understand the dynamics of how NLP models make mistakes on unusual inputs, in order to improve accuracy, increase robustness, and maintain security or privacy. While this line of research has recently received a lot of attention in the deep learning community, it has a long history in machine learning, going back to adversarial attacks on linear spam classifiers \cite{dalvi2004adversarial,lowd2005adversarial}.
% As a real-world application, 
%Hosseini et al. \shortcite{hosseini2017deceiving} observed that simple modifications, such as adding spaces or dots between characters,  could sometimes drastically change the toxicity score from Google's \texttt{perspective} API. 

Character-level NMT systems \cite{lee2016fully,costa2016character} and those based on sub-word units \cite{sennrich2015neural} are able to extract morphological features which can generalize unseen words.  
%f these filters are easy to evade, then they will be ineffective in practice, even if they perform well on held-out data. 
%By identifying weaknesses of the models before deployment, we can select models that will be harder to attack.  
% In one of the first attempts at tricking deep neural classifiers, Papernot et al., \shortcite{papernot2016crafting} added adversarial noise to the word embeddings in an LSTM and searched their neighborhood to find a word to replace the original word. While their adversary was able to trick the classifier, their word-level changes do not preserve the meaning (e.g., ``I wouldn't rent this...'' $\rightarrow$ ``Excellent wouldn't rent this...'').
  Belinkov and Bisk \shortcite{belinkov2017synthetic} show that character-level machine translation systems are overly sensitive to random character manipulations, such as keyboard typos. They use black-box heuristics to generate character-level adversarial examples, without using the model parameters or gradients to generate adversarial examples. 
  The major challenge in creating white-box adversarial examples for text is that optimizing over discrete input is difficult \cite{miyato2016adversarial}, which is why previous work has focused on black-box adversarial examples. Zhao et al. \shortcite{zhao2017generating} search for black-box adversarial examples in the space of encoded sentences and generate adversarial examples by perturbing the latent representation until the model is tricked. However, it is not clear how many queries are sent to the model or what the success rate of the adversary is. We contrast black-box and white-box attacks and show how white-box attacks significantly outperform their black-box counterparts in controlled and targeted attack scenarios. 

Adversarial training/regularization interleaves training with generation of adversarial examples \cite{goodfellow2014explaining}. Concretely, after every iteration of training, adversarial examples are created and added to the mini-batches.
% A projected gradient-based approach to create adversarial examples by Madry et al. \shortcite{madry2017towards} has proved to be one of the most effective defense mechanisms against adversarial attacks for image classification.
%%Virtual adversarial training \cite{miyato2015distributional} is another regularization method, which aims to minimize the KL divergence between predictions on the examples and their adversarial counterparts. 
This technique has been used for text classification \cite{miyato2016adversarial} using adversarial noise on the word embeddings without creating real-world adversarial examples. Jia and Liang \shortcite{jia2017adversarial} point out the difficulty of adversarial training with real-world adversarial examples, as it is not easy to create such examples efficiently. 
%, but is has focused solely on improving accuracy on clean examples, without creating textual adversarial examples.
 In this work, we improve the training time of HotFlip \cite{ebrahimi2017hotflip} by using a one-shot attack in our inner adversary, which makes the running time of adversarial training only 3 times slower than regular training.

\section{White-Box Adversarial Examples}
%We devise a beam search optimization method which applies gradient-driven perturbations to the discrete input. 
Editing text to trick an NLP model, constrained by the number of  characters to change, $r$, is a combinatorial search problem. We develop a gradient-based optimization method to perform four text edit operations: namely, flip (replacing one character with another), swap (replacing two adjacent characters with each other), delete, and insert.
%Our method is based on an atomic flip operation, which can be extended to insertion, deletion, and swap operations, in order to constitute a more comprehensive set of adversarial attacks to trick an NMT system. 
We use derivatives with respect to one-hot representation of the input, to rank candidate changes to text, in order to search for an adversarial example which satisfies the adversary's goal. Compared with a black-box adversary, our method has the overhead of sorting or searching among derivatives, while being considerably more successful. 

%Specifically, we used a CharCNN-LSTM encoder and a word-based attentional decoder \cite{bahdanau2014neural}. 
%This method achieves the goal of fooling the target system by applying perturbations that increase the loss of the model. In addition, when a model employs HotFlip during training, it will become more robust against attacks at test time.
%The architecture we study is based on the one proposed by Kim et al. \shortcite{kim2015character} for character-level language modeling. 
%Feature extraction is performed by convolutions over characters, which are passed to layers of highway networks, and finally given to stacks of recurrent neural nets for modeling a sequence of words.
%Our method of generating adversarial examples can be used with any character-level architecture.
%The reason that we pick this architecture is its flexibility to accommodate several NLP tasks; for example,
%This architecture has been used to perform sequence labeling \cite{kim2015character} and machine translation \cite{costa2016character,belinkov2017synthetic}. 
%We adapt it for our text classification experiments, wherein the output of the last recurrent unit is passed to a softmax to predict the label of text. 
%For our word-level text classification experiments, we use Kim's convolutional neural model \shortcite{kim2014convolutional} which has become a standard baseline for neural text classification.
%See Figure~ \ref{arch}. 
%We refer the reader to  Kim et al. \shortcite{kim2015character} for more details.
\subsection{Definitions}
%We use character-level text classification as our running example, which is easily generalizable to a word-level model.
We use $J(\mathbf{x}, \mathbf{y})$ to refer to the log-loss of the translation model on source sequence $\mathbf{x}$ and target sequence $\mathbf{y}$. 
%For example, for classification, the loss would be the log-loss over the output of the softmax unit.
Let $V$ be the alphabet, $\mathbf{x}$ be a text of length $L$ characters, and $x_{ij} \in \{0,1\}^{|V|}$ denote a one-hot vector representing the $j$-th character of the $i$-th word. The character sequence can be represented by
\centerline{$\mathbf{x}$ = [($x_{11}$,.. $x_{1n}$);($x_{21}$,.. $x_{2n}$);...($x_{m1}$,.. $x_{mn}$)]}
wherein a semicolon denotes explicit segmentation between words. The number of words is denoted by $m$, and $n$ is the number of maximum characters allowed for a word\footnote{Padding is applied if the number of characters is fewer than the maximum.}. 
%For word-based representations, we instead have $x_i  \in \{0,1\}^{|V|}$ as a one-hot vector representing the $\text{i}_\text{th}$ word, where $V$ is the vocabulary. Either of these representations are passed to a corresponding embedding layer, which is then fed to the network. In the following description of our method, we use the character-based representation as our running example. 

%Characters in text are analogous to pixel intensities in images. Unlike continuous data, discrete data cannot be infinitesimally changed. We employ three basic string operations to change text: flipping, insertion, and deletion of a single character.
%It is true that the meaning of a word depends on its orthography \cite{de1916nature}, and these string operations are noticeable changes. By limiting the number of these manipulations to a just a few, we can achieve the smallest possible changes to the input and also preserve the meaning of the text.
%In the next subsection, we describe our approach in detail using the notation in the previous section, and classification as our running example.
\subsection{Derivatives of Operations}
%We describe three basic string operations that would be used to trick different NLP systems, namely flipping, insertion, and deletion.
%Let $J(\mathbf{x}, \mathbf{y})$ denote the log-loss over the output of the softmax unit. 
 % We abbreviate $J(\mathbf{x}, \mathbf{y})$ as $J(\mathbf{x})$ or J when the meaning is clear from context.
%Imagine the adversary is allowed to change $r$ characters in the input text to fool the model to which it has access. 
%Using a brute-force search, it would need to do $\binom{L}{r}|V|^r$ forward passes to exhaust the search space. That is, query the classifier, by calling $J$, for all combinations of character flips within the allowed budget, $r$. This can decrease to $\mathcal{O}(brL|V|)$ if beam search is used. That is, after trying all possible single character flips, keep the top $b$ flips which had the highest loss increase, and continue for $r$ steps.
We represent text edit operations as vectors in the input space, and estimate the change in loss by directional derivatives with respect to these operations. Based on these derivatives, the adversary can choose the best loss-increasing operation. Our algorithm requires just one function evaluation (forward pass) and one gradient computation (backward pass) to estimate the best possible flip. 

A  \textbf{flip} of the $j$-th character of the $i$-th word ($a \rightarrow b$) can be represented by this vector:
 
\centerline{
$\vec{v}^{\,}_{ijb}$ = ($\vec{0}^{\,}$,..;$($$\vec{0}^{\,}$,..$($0,..-1,0,..,1,0$)_j$,..$\vec{0}^{\,}$$)_i$; $\vec{0}^{\,}$,..)
} 
where -1 and 1 are in the corresponding positions for the $a$-th and $b$-th characters of the alphabet, respectively, and $x_{ij}^{(a)}=1$. A first-order approximation of change in loss can be obtained from a directional derivative along this vector: %Due to directional derivatives, we have the following:
 
 \begin{equation}
 \nabla_{\vec{v}^{\,}_{ijb}}J(\mathbf{x}, \mathbf{y}) = \nabla_{x}J(\mathbf{x}, \mathbf{y})^{T} \cdot \> \vec{v}^{\,}_{ijb} = {{\frac{\partial{J}}{\partial{x_{ij}}}}^{(b)} - { \frac{\partial{J}}{\partial{x_{ij}}}}^{(a)}}    
 \label{eq1}
 \end{equation}
 
%We choose the vector with biggest increase in loss:
%%The vector with the HotFlip direction, among all possible character flips can be easily found:
%\begin{equation}
%\begin{split}
%%\lVert{\vec{v_{ij}}^{\,}}\rVert = 1
%%\underset{}{\text{max}} \nabla_{x}J(\mathbf{x}, \mathbf{y})^{T} \cdot \> \vec{v}^{\,}_{ijb}= \underset{\vec{v}^{\,}_{ijb}}{\text{max}} {{\frac{\partial{J}}{\partial{x_{ij}}}}^{(a)} - { \frac{\partial{J}}{\partial{x_{ij}}}}^{(b)}}
%\underset{}{\text{max}} \nabla_{x}J(\mathbf{x}, \mathbf{y})^{T} \cdot \> \vec{v}^{\,}_{ijb}= \underset{{ijb}}{\text{max}} {{\frac{\partial{J}}{\partial{x_{ij}}}}^{(b)} - { \frac{\partial{J}}{\partial{x_{ij}}}}^{(a)}}
%\end{split}
%\label{eq1}
%\end{equation}

%An immediate benefit of using derivatives with respect to the one-hot vectors is that they can be used to \textit{esimate} the best character change ($a \rightarrow b$). 
%Concretely, the derivative vector contains the information about the loss increase obtained due to a character flip. 
%Applying the resulting vector to the input amounts to the smallest gradient ascent step in the input space. 
Using the derivatives as a surrogate loss, we simply need to find the best change by \textbf{maximizing} eq. \ref{eq1}, to \textit{estimate} the best character change ($a \rightarrow b$). This requires searching in $|V|mn$ values for a given text of $m$ words with $n$ characters each, in a vocabulary of size $|V|$. 
%Using the derivatives, we simply need to find the best change by calling the function mentioned in eq. \ref{eq1}, to \textit{esimate} the best character change(s) ($a \rightarrow b$).  This is in contrast to a naive loss-based approach, which has to query the classifier for every possible change to compute the \textit{exact} loss induced by those changes. In other words, one backward pass saves us a large number of forward passes.
%e number of queries be independent of the alphabet size, which is crucial for word-level models with considerably larger alphabet size than character-level models. 

%Character \textbf{insertion}\footnote{For ease in exposition, we assume that the word size is at most $n$-1, leaving at least one position of padding at the end.} at the $j$-th position of the $i$-th word can also be treated as a character flip, followed by more flips as characters are shifted to the right until the end of the word. Hence the estimate in change of loss corresponding to a character insertion can be given by: 
%\begin{equation}
%% \begin{split}
% {{\frac{\partial{J}}{\partial{x_{ij}}}}^{(b)}-{\frac{\partial{J}}{\partial{x_{ij}}}}^{(a)}}  
%+ \sum_{j^{'}=j+1}^n \bigg( {{\frac{\partial{J}}{\partial{x_{ij^{'}}}}}^{(b^{'})} - { \frac{\partial{J}}{\partial{x_{ij^{'}}}}}^{(a^{'})}} \bigg)
%%\end{split}
%\label{eq-der}
%\end{equation}
%where $x_{ij^{'}}^{(a^{'})}=1$ and $x_{i{j^{'}{-1}}}^{(b^{'})}=1$.

Similarly, character \textbf{insertion}, \textbf{deletion}, and \textbf{swap} of adjacent characters can be represented as vectors which carry the information about the direction and number of flips. 
%The maximizing direction vector is:
%\begin{equation}
%% \begin{split}
%%\underset{}{\text{max}} \nabla_{x}J(\mathbf{x}, \mathbf{y})^{T} \cdot \> \vec{v}^{\,}_{ijb}=
%\underset{{ij}}{\text{max}} \sum_{j^{'}=j}^{n-1} \bigg( {{\frac{\partial{J}}{\partial{x_{ij^{'}}}}}^{(b^{'})}-{ \frac{\partial{J}}{\partial{x_{ij^{'}}}}}^{(a^{'})}} \bigg)
%%\end{split}
%\end{equation}
%where $x_{ij^{'}}^{(a^{'})}=1$ and $x_{i{j^{'}{+1}}}^{(b^{'})}=1$.
Since the magnitudes of operation vectors are different, we normalized them by both $L_1$ and $L_2$ norm but found it had little impact.

A black-box adversary can perform similar manipulations which would be randomly picked. For example, Belinkov and Bisk \shortcite{belinkov2017synthetic} define \texttt{Key} operation, which flips one character with an adjacent one on the keyboard, at random. 
%\subsection{Controlled Attack}
%The previous adversary was untargeted and its only goal was to increase the loss of the model. However, a controlled attack on an NMT will focus on word of the target sequence and mask the loss for other words in the sequence. Concretely, we use $J(\mathbf{x}, \mathbf{y}_t)$ as our loss function, where $t$ is the target word. This way, the adversary can focus on the target word and thus change parts of input that would effect that target word most.
%%These normalized vectors can further be scaled by some constant $\alpha$ based on the effectiveness of the operation, or other criteria that the adversary prefers.
%%As our experiments show, the most effective change is the flip.
%
%\subsection{Targeted Attack}
%A more difficult attack is when the adversary aims to not only mute a word but also change it to another word.
%%Our adversary can also inflict targeted attacks, in which instead of simply changing the output, it pursues the goal of changing the output to a desired target output. 
%As the following equation denotes, the adversary should try to \textbf{minimize} the estimated error, given the new target word $\mathbf{y_{t^{'}}}$.
%\begin{equation*}
%\begin{split}
%\underset{}{\text{min}} \nabla_{x}J(\mathbf{x}, \mathbf{y_{t^{'}}})^{T} \cdot \> \vec{v}^{\,}_{ijb}= \underset{{ijb}}{\text{min}} {{\frac{\partial{J}}{\partial{x_{ij}}}}^{(b)} - { \frac{\partial{J}}{\partial{x_{ij}}}}^{(a)}}
%\end{split}
%\end{equation*}
 
 \subsection{Controlled and Targeted Attacks}
 The previous adversary was untargeted, and its only goal was to increase the loss of the model. However, some corruptions of the output may be much worse than others -- translating ``good morning'' as ``attack them'' is much worse than translating it as ``fish bicycle.'' By changing the loss function, we can force the adversary to focus on more specific goals. 

In a \emph{controlled attack}, the adversary tries to remove a specific word from the translation. This could be used to maintain privacy, by making more sensitive information harder to translate, or to corrupt meaning, by removing key modifiers like ``not,'' ``joked,''' or ``kidding.'' Concretely, we maximize the loss function $J(x, y_t)$, where $t$ is the target word. This way, the adversary ignores the rest of the output and focuses on parts of the input that would affect the target word most.

In a \emph{targeted attack}, the adversary aims to not only mute a word but also replace it with another. For example, changing the translation from ``good morning''' to ``good attack'' could lead to an investigation or an arrest. Making specific changes like this is much more dangerous, but also harder for the adversary to do. For this attack, we maximize the loss $-J(x, y_{t^{'}})$, where $t{^{'}}$ is the new word chosen to replace $t$.  Note that the negation makes this equivalent to minimizing the predictive loss $J(x, y_{t^{'}})$ on $t^{'}$. We represent this as maximization so that it fits in the same framework as the other attacks.
Our derivative-based approach from the previous subsection can be then used directly to generate these new attacks, simply by substituting the alternate loss function.
 
%A similar beam search strategy is required to search the input space until the current output is changed to the target output or the adversary is run out of budget.

 %For this experiment, we considered only flip changes.
%the output of the original input, than a random direction on the same word on which the best direction occurs.
%For this experiment, we considered only flip changes, and computed the average squared distance between the output of the original input and the modified ones after the CNN and highway layers.

 \subsection{Multiple Changes}
We explained how to estimate the best single change in text to get the maximum increase/decrease in loss. We now discuss approaches to perform multiple changes.
\begin{enumerate}[label=(\alph*)]
\item \textbf{one-shot:} In this type of attack, the adversary manipulates all the words in the text with the best operation in parallel. That is, the best operation for each word is picked locally and independently of other words. This is efficient, as with only one forward and backward pass, we can collect the gradients for all operations for all words. In addition, compared with the global one-shot method of Ebrahimi et al. \shortcite{ebrahimi2017hotflip}, it does not require sorting the gradients globally, and can further reduce the time to create adversarial examples. It is less optimal than the next approaches, which apply changes one by one. Due to its efficiency, this is the approach we choose to do adversarial training. Our experiments in section \ref{wb:num1}, which are untargeted, follow this approach. We also investigate black-box and white-box variants of one-shot attacks in section \ref{u:num1}. The budget for the adversary is the number of words, and it is spent in the first shot. 
\item \textbf{Greedy:} In this type of attack, after picking the best operation in the whole text, we make another forward and backward pass, and continue our search. Our controlled attack in section \ref{ma:num1} follows this approach, where we allow a maximum of 20\% of the characters in text as the budget for the adversary. As will be explained in \ref{ma:num1}, the adversary spends much less than this amount.
\item \textbf{Beam Search}: And finally, we can strengthen our greedy search by beam search. Our beam search requires only $\mathcal{O}(br)$ forward passes and an equal number of backward passes, with $r$ being the budget and $b$, the beam width. At every step, the beam will be sorted by the sum of the true loss up to that point, which we have computed, plus the gradient-based estimate of candidate operations. This showed better performance than using the sum of the gradients in the path for sorting the beam, as was previously done \cite{ebrahimi2017hotflip}. Since targeted attacks are the most difficult type of attack, we use this strategy for targeted attacks, as described in section \ref{ta:num2}. We allow a maximum of 20\% of the characters in text as the budget for the adversary, and set the beam width to 5.
\end{enumerate}

\section{Experiments}
%conduct a broad set of experiments for two tasks and 
%analyze adversarial examples for character-level translation. Our seq-2-seq implementation relies largely on Yoon Kim's seq2seq implementation \footnote{https://github.com/harvardnlp/seq2seq-attn}, which mostly follows the guidelines of Luong et al. \shortcite{luong2015effective} for attentional translation. Specifically, we used a CharCNN-LSTM encoder and a word-based attentional decoder \cite{bahdanau2014neural}. Unless otherwise stated, we use a beam size of 4 for decoding at test time. 
We use the TED talks parallel corpus prepared by IWSLT 2016 \cite{mauro2016iwslt} for three pairs of languages: German to English, Czech to English, and French to English. We use the development sets and test sets of previous years except 2015 as our development set. The statistics of the dataset can be found in Table \ref{stats}.
%Preprocessing consisted of tokenizing, normalizing punctuation, and filtering sentences with more than 50 words. 
%For text classification, we use the {AG's news} dataset\footnote{https://www.di.unipi.it/\textasciitilde gulli/}, which consists of 120,000 training and 7,600 test instances from four equal-sized classes: World, Sports, Business, and Science/Technology.
%\footnote{https://www.di.unipi.it/\textasciitilde gulli/AG\_corpus\_of\_news\_articles.html}
%The architecture consists of a 2-layer stacked LSTM with 500 hidden units, and a character embedding size of 25, and we used the default values for hyper-parameters 
%For classification, we used 10\% of the training data as the development set, and trained for a maximum of 25 epochs. 
%For translation, we use OpenNMT's suggested hyper-parameters.
Throughout our experiments, we only allow character changes if the new word does not exist in the vocabulary, to avoid changes that an MT would respond to as expected. For example, it is not surprising that changing the source German word ``nacht'' to ``nackt'' would cause an MT to introduce the word ``nude'' in the translation. 

The architecture we study was first proposed by Kim et al. \shortcite{kim2015character} for language modeling, and later adapted by Costa-Jussa and Fonollosa \shortcite{costa2016character} for translation, and is also used by Belinkov and Bisk \shortcite{belinkov2017synthetic} for their adversarially-trained models. In this architecture, feature extraction is performed by convolutions over characters, which are passed to layers of highway networks, and finally given to stacks of recurrent neural nets for modeling a sequence of words. Using this architecture, the BLEU scores on our datasets are competitive with the submissions to the IWSLT \cite{mauro2016iwslt}.  Our implementation\footnote{https://github.com/jebivid/adversarial-nmt} relies largely on Yoon Kim's seq2seq implementation\footnote{https://github.com/harvardnlp/seq2seq-attn}, with similar hyper-parameters, which mostly follows the guidelines of Luong et al. \shortcite{luong2015effective} for attentional translation. 

For experiments in section \ref{wb:num1}, where we report the BLEU score for a vanilla model and several adversarially-trained models against different attackers, we use a decoder with a beam width of 4. However, our white-box attacker uses a model with greedy decoding to compute gradients. The reason is that in order to calculate correct gradients, we either need to use greedy decoding, or use models which incorporate beam search in the decoder architecture such that gradients could flow in the beam paths, too \cite{wiseman2016sequence}, which incur more computational cost. Correct gradients are more of an important issue for targeted attacks, where we want to achieve a goal beyond simply breaking the system. For the sake of consistency, we use greedy decoding for both the vanilla model which is being attacked, and the white-box attacker in all experiments of section \ref{greedydecoding}, where we contrast white-box and black-box adversaries in different scenarios.
\begin{table}
\centering{
\scalebox {0.9} {
\begin{tabular}{c|c|c|c  }
Pair &  Train & Test & Target vocab.  \\
% \hline
%misc. error \% &83.16 & 69.38& 66.32 & 81.97 \\
\hline
FR-EN & 235K & 1.1k & 69k \\
\hline
DE-EN  & 210K & 1.1k & 66k   \\
\hline
CS-EN  & 122K & 1.1k &   49k    \\
\hline
\end{tabular}
}
}
 \captionsetup{font=footnotesize}
 \caption{Data Statistics }
 \label{stats}
\end{table}

\section{Analysis of Adversaries} \label{greedydecoding}
\begin{figure}
\centering
\scalebox{1}{
\includegraphics[width=\linewidth]{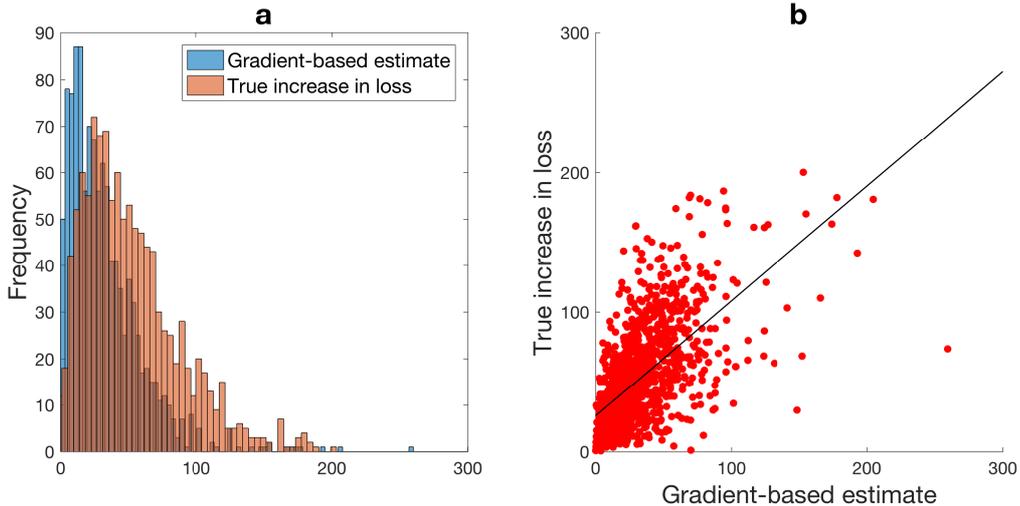}
}
\captionsetup{font=footnotesize}
\caption{Comparing the distribution of the true increase in loss and its gradient-based estimate, and their correlation, using best flips for each word in a sentence of the German test set.}\label{losshist}
\end{figure}

We first study whether first-order approximation gives us a good estimator to be employed by white-box adversaries. Figure \ref{losshist} compares the true increase in log-loss (i.e., $J(x+v) - J(x)$), with our gradient-based estimate (i.e.,  $\nabla_{v}J(x)$). We create adversarial examples using the best estimated character flip for every word, over the German test set. Then, we compare the true increase in loss for the created adversarial examples, with our gradient-based estimate. The log-loss is evaluated by summing the log-loss of individual words, and similarly, the gradient-based estimate is the sum of all gradients given by flips which are performed once on every word. Figure \ref{losshist}.a plots the histograms for both of these measures, and Figure \ref{losshist}.b shows a scatter plot of them with the least squares fitting line. Due to linearization bias of the first-order approximation, we have a distribution with smaller variance for the gradient-based estimate measure. We can also observe a moderately positive correlation between the two measures (Spearman coefficient $\rho=0.61$), which shows we can use the gradient-based estimate for \textit{ranking} adversarial manipulations.

Next,  we contrast black-box and white-box adversaries in untargeted, controlled, and targeted scenarios, and demonstrate that white-box adversaries significantly outperform black-box adversaries especially in controlled and targeted scenarios.
 \subsection{Untargeted Attack} \label{u:num1}
 Table \ref{bleuadv} shows the BLEU score after one-shot white-box and black-box attacks are performed. 
 %The results for the black-box adversaries are averaged over 10 runs. 
Unlike delete and swap, insert and flip have the advantage of making changes to one-letter words, so we expect them to perform better.
%, we disallow changing them for all types adversaries for the sake of fair comparison.
We see this for the white-box attacks which can pick the best change to every word using the gradients. On the contrary, a black-box adversary performs worst for flip, which is because the black-box attacker is not enabled to pick the best change when more options (possible character flips) are available, as opposed to swap and delete which are governed by the location of the change and contain no additional flip. Nevertheless, a black-box adversary has competitive performance with the white-box one, even though it is simply randomly manipulating words. We argue that evaluating adversaries, based on their performance in an untargeted setting on a brittle system, such as NMT, is not appropriate, and instead suggest using goal-based attacks for evaluation.

%As can be seen, white-box adversaries are stronger than their black-box counterparts. This proves that a
%white-box adversary, which has access to model parameters, can use a gradient-based method to select
%more damaging operations. 
%White-box flip operations are much more effective than their black-box counterparts.
%In contrast, black-box delete operation is quite effective and decreases the BLEU score almost
%as well as the white-box delete. This suggests that single black-box character flips are less effective than
%black-box operations which include more flips, such as delete. However, the white-box adversary can be
%successful even with single flips. 

%Since we are not optimizing to decrease the BLEU score directly, we
%also include the log-loss values in Table \ref{loglossadv}, which shows the effectiveness of the white-box adversary in
%increasing the loss.
  \begin{table*}[ht]
   \centering
 \scalebox{1}{
  \begin{tabular}{c|c|c||c|c||c|c||c|c|}

% {Attack} &\multicolumn{2}{c|}{Flip}  &\multicolumn{2}{c|}{Insert} &\multicolumn{2}{c|}{Delete} &\multicolumn{2}{c|}{Swap}\\  
%\hline
%Source  & white & black  & white & black & white & black & white & black\\
% 
 
 \multirow{2}{*}{Attack} &\multicolumn{2}{c}{Flip}  &\multicolumn{2}{c}{Insert} &\multicolumn{2}{c}{Delete} &\multicolumn{2}{c}{Swap} 
 
\\
% \hhline{~--------}
%    & white & black  & white & black & white & black & white & black\\
%    \hline
%         CS & \underline{\textbf {5.10}} &  {7.56} & \textbf {5.40} &  {7.27} & \textbf {7.86} &  {7.90} & \textbf {7.08} & 8.55  \\
%    \hline
%      DE & \textbf {5.02} &  {7.74} & \underline{\textbf {4.41}} &  {4.96} & {6.66} &   \textbf{6.53} & \textbf {5.63} & 6.15  \\
%    \hline
%    FR & \underline{\textbf {4.98}} &  {8.81} & \textbf {5.45} &  {5.90} & \textbf {6.12} &  {6.79} & \textbf {5.72} &  6.81  \\
%    \hline
  
 \hhline{~--------}
 
    & white & black  & white & black & white & black & white & black\\
    \hline
        FR & { \underline{\textbf {4.27}}} &  {6.98} & \textbf {4.74} &  {4.85} & \textbf {4.99} &  {5.86} &  {\textbf {4.87}} &  5.20  \\
    \hline
          DE & {\textbf{4.50}} &  {6.87} & \underline{\textbf {3.91}} &  {4.31} &  \textbf{5.63} &   {5.73} & {4.94} & \textbf{4.74}  \\
    \hline
         CS & \underline{\textbf {4.31}} &  {6.09} & \textbf {4.66} &  {5.86} & \textbf {6.30} &  {6.62} &   {6.05} & \textbf{5.82}  \\
    \hline

  \end{tabular}
    }
  \caption{BLEU score after greedy decoding in the existence of different types of untargeted attacks.}  
  \label{bleuadv}
\end{table*}

\subsection{Controlled Attack}\label{ma:num1}
We introduce more interesting attacks, in which the adversary targets the MT for more specific goals. A  {perfect} \textit{mute} attack removes a word successfully and keeps the rest of the sentence intact. For example, consider the translation $T$, containing words $w_1$, $w_2$,..., $w_t$, ...$w_n$, where $w_t$ is the target word. A {perfect} mute attack will cause the NMT to create a translation $T_p$ which contains words $w_1$, $w_2$,..., UNK, ...$w_n$, wherein $w_t$ is replaced with UNK. With this observation in mind, we define the success rate of an attack, which generates $T_{adv}$, as follows:

\[
    \text{success}(T_{adv})= 
\begin{cases}
    1 ,& \text{if } \frac{\text{BLEU}(T, \>T_{adv})}{\text{BLEU}(T, \>T_{p})} \geq \alpha \\
    0,              & \text{otherwise}
\end{cases}
\]

\begin{figure}
\centering
\scalebox{0.95}{
\includegraphics[width=\linewidth]{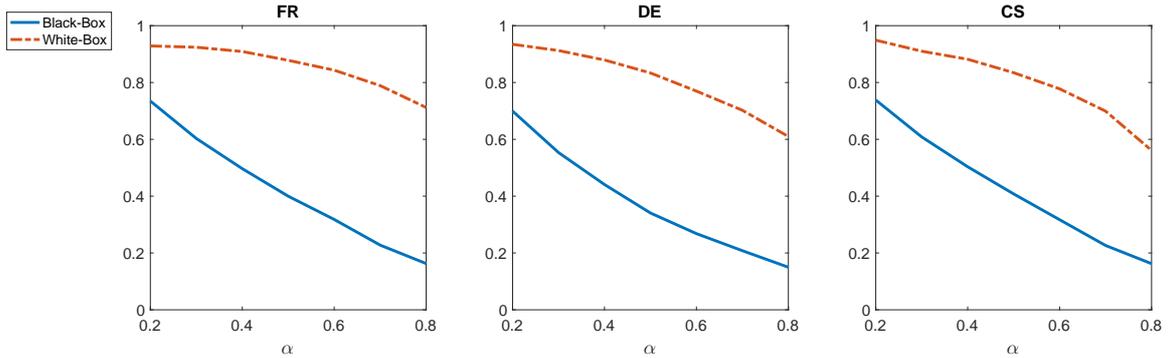}
}
\captionsetup{font=footnotesize}

\caption{Success rate of white-box and black-box adversaries in a controlled setting as a function of $\alpha$.}\label{fig:successadversary}
\end{figure}

We can control the quality of an attack with $\alpha$, for which a larger value punishes the adversary for ad-hoc manipulations, which could cause the NMT to generate a radically different and possibly gibberish translation. Success rate is defined by the number of successful attacks divided by the number of total sentences in the test set. Figure \ref{fig:successadversary} plots the success rate against $\alpha$. As can be seen, the white-box adversary is significantly more successful than a black-box adversary. By taking advantage of the knowledge of gradients of the model, a white-box adversary can perform better targeted attacks.  For this experiment, the black-box attacker uniformly picks from the four possible changes and randomly applies them. 

For this attack, we follow the greedy approach, and use a budget of 20\% of the characters in text. Table \ref{efficiency} shows the average number of character changes and the number of queries made to the model. The reported character changes are for attacks wherein the attacker only muted a target word successfully, regardless of the quality of the translation. The reported number of queries takes the unsuccessful trials into account too. The white-box adversary is more efficient due to fewer queries and fewer manipulated characters, which can be crucial for a real-word adversary. Nevertheless, unlike a black-box adversary, a white-box adversary requires additional backward passes, and has the overhead of operations on the gradient values, mainly sorting. This makes the running times of the two comparable.

Compared with the results in the previous section, controlled attacks show a more convincing superiority of the white-box attacks over black-box attacks. 

%\begin{equation}
% \text{success}_\alpha = \frac{\mid \frac{\text{BLEU}(T, T_{adv})}{\text{BLEU}(T, T_{p})} \ge \alpha \mid}{|D|}
%\end{equation}

 \begin{table*}[ht]
   \centering
 \scalebox{0.8}{
  \begin{tabular}{c c c||c c}

% 
% \multirow{2}{*}{} &\multicolumn{2}{c}{Character Changes}  &\multicolumn{2}{c}{Queries}  
% 
%\\ 
% \hhline{~----}

 {} &\multicolumn{2}{c}{Character Changes}  &\multicolumn{2}{c}{Queries}  \\ 
  \hline
  
   source    & white & black  & white & black \\
       \hline
    FR & {\textbf {1.9}} &  {7.7} & \textbf {2.3k} &  {8.9k}   \\

    \hline
      DE & {\textbf {1.9}} &  {6.5} & {\textbf {1.9k}} &  {7.8k}   \\

       \hline
         CS &  {\textbf {1.5}} &  {5.3} & \textbf {1.2k} &  {6.1k}  \\

  \end{tabular}
    }
  \caption{Efficiency of attacks.}  
  \label{efficiency}
\end{table*}

\subsection{Targeted Attack}\label{ta:num2}
 A more challenging attack is to not only mute a word but also replace it with another one. The evaluation metric for targeted attack is similar to controlled attack with one difference that a perfect \textit{push} attack produces a translation, $T_p$, which contains words $w_1$, $w_2$, $w_{t^{'}}$,..$w_n$, wherein $w_{t^{'}}$ has replaced $w_{t}$. In classification domains with few classes, targeted attacks are relatively simple, since an adversary can perturb the input to move it to the other side of a decision boundary. Whereas in MT, we deal with vocabulary sizes in the order of at least tens of thousands, and it is less likely for an adversary to be successful in targeted attacks for most possible target words.  To address this, we evaluate our adversary with $n^\text{th}$-most likely class attacks. In the simplest case, we replace a target word with the second-most likely word at decoding time. 
 
\begin{figure}
\centering
\scalebox{0.95}{
\includegraphics[width=\linewidth]{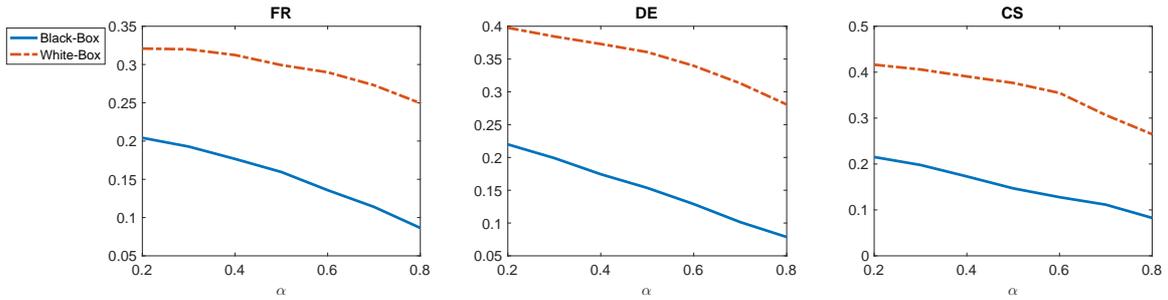}
}
\captionsetup{font=footnotesize}

\caption{Success rate of white-box and black-box adversaries in the second-most likely targeted attack as a function of $\alpha$.}\label{fig:successadversary2}
\end{figure}

 As can be seen in Table \ref{fig:successadversary2}, targeted attacks are much more difficult with a much lower success rate for the adversary. Nevertheless, the white-box adversary still performs significantly better than the black-box adversary. The success rate dramatically goes down for large values of $n$. For example, for the value of 100, the success rate will be more than ten times smaller than second-most likely attack.
 
\subsection{Some Adversarial Examples}
Table \ref{exmp-trans-targfull} shows three adversarial examples. The first example shows a controlled attack, where the adversary has successfully removed a swear word from the sentence. The BLEU ratio, used in our success rate measure, for this example is 0.52. The second example shows a second-most likely targeted attack where the new translation has managed to keep the rest of the translation intact and achieve its goal. The BLEU ratio for this example is 1.00. The third example, which has a BLEU ratio of 0.70, shows a $100^\text{th}$-most likely attack, where the word \textit{competition} is replaced with \textit{throwaway}. Due to the difficulty of this change, the adversary has committed a considerably larger number of manipulations.
\begin{table*}
% \vspace{-20pt}
\centering
{\renewcommand{\arraystretch}{0.9}
\scalebox{0.95}{
\begin{tabular}{p{1.4cm}p{14cm}}
\hline
 \small{src} & \footnotesize{ Wir erwarten Perfektion von Feministinnen, weil wir immer noch für so viel k{\"a}mpfen, so viel wollen, so verdammt viel brauchen.}\\
\small{adv} & \footnotesize{Wir erwarten Perfektion von Feministinnen, weil wir immer noch für so viel k{\"a}mpfen, so viel wollen, so \textbf{öberdammt} viel brauchen.}\\
\small{src-output} & \footnotesize{We expect perfection from feminist, because we're still fighting so much, so \textbf{damn}, so \textbf{damn}, so \textbf{damn}.}\\
\small{adv-output} & \footnotesize{We expect perfection from feminist, because we still fight for so much, so much of all, so much of all that needs to be.}\\
 \hline
  
   \small{src} & \footnotesize{In den letzten Jahren hat sie sich zu einer sichtbaren Feministin entwickelt.}\\
\small{adv} & \footnotesize{In den letzten Jahren hat sie sich zu einer sichtbaren \textbf{FbeminisMin} entwickelt.}\\
\small{src-output} & \footnotesize{In the last few years, they've evolved to a safe \textbf{feminist}.}\\
\small{adv-output} & \footnotesize{ In the last few years, they've evolved to a safe \textbf{ruin}.}\\

   \hline
   
      \small{src} & \footnotesize{Ein Krieg ist nicht l{\"a}nger ein Wettbewerb zwischen Staaten, so wie es früher war.}\\
\small{adv} & \footnotesize{Ein Krieg ist nicht l{\"a}nger ein \textbf{erkBkaSzeKLlWmrt} zwischen Staaten, so wie es früher war.}\\
\small{src-output} & \footnotesize{A war is no longer a \textbf{competition} between states, like it used to be.}\\
\small{adv-output} & \footnotesize{A war is no longer a \textbf{throwaway} planet between states, as it used to be.}\\

   \hline

\end{tabular}
}
}
%\vspace{-20pt}
\captionsetup{font=footnotesize}
\caption{A controlled attack and two targeted attacks on our DE-EN NMT. First example shows a controlled attack, the second and third examples show a second-most and a $\text{100}_\text{th}$-most likely targeted attack, respectively.}
\label{exmp-trans-targfull}
\end{table*}

\section{Robustness to Adversarial Examples}  \label{wb:num1}
\subsection{Baselines}
We use the black-box training method of Belinkov and Bisk \shortcite{belinkov2017synthetic} as our baseline. They train several models using inputs which include noise from different sources. We used their script\footnote{https://github.com/ybisk/charNMT-noise} to generate random (\texttt{Rand}), keyboard (\texttt{Key}), and natural (\texttt{Nat}) noises. Their best model was one which incorporated noise from all three sources (\texttt{Rand+Key+Nat}). 

\begin{figure}
\centering
\scalebox{0.6}{
\includegraphics[width=\linewidth]{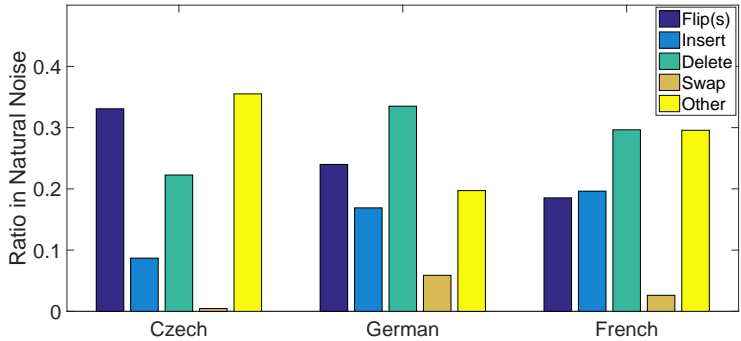}
}
\captionsetup{font=footnotesize}

\caption{Distribution of types of noise in the natural noise corpora.}\label{naturalproportion}
\end{figure}

Similar to their approach, we train a model which incorporates noisy input scrambled by Flips, Inserts, Deletes, and Swaps in training (\texttt{FIDS-B}). Since natural noise was shown to be the most elusive adversarial manipulation \cite{belinkov2017synthetic}, we used this source of noise to determine the proportion of each of the FIDS operations in training. Concretely we found that the majority of the natural noise can be generated by FIDS operations, and we used the ratio of each noise in the corpora to sample from these four operations. 
Figure \ref{naturalproportion} shows the distribution of manipulations for each language. 
A single swap is the least likely operation in all three languages\footnote{Excluding single swaps from manipulations with two flips.}. 
FIDS operations account for 64\%, 80\%, and 70\% of natural noise for Czech, German, and French, respectively. This can be regarded as a background knowledge incorporated into the adversary.

 \subsection{White-Box and Ensemble Methods}

Our white-box adversary \texttt{FIDS-W} generates adversarial examples using our four text edit operations in accordance with the distribution of operations on the natural noise. At every epoch, adversarial examples are generated for every mini-batch using the one-shot approach. More precisely, all words in the sentence are changed by a single FIDS operation in parallel. While training on both clean and adversarial examples has been the standard approach in adversarial training, some evidence in computer vision \cite{madry2017towards,shaham2015understanding} suggests that training on white-box adversarial examples alone can boost models' robustness to adversarial examples, with a minor decrease in accuracy on clean examples. However, we found that in order to get a good BLEU score on the clean dataset, we need to train on both clean and white-box adversarial examples. 
We also train an ensemble model, \texttt{Ensemble}, which incorporates white-box and black-box adversarial examples, with 50/50 share for each. The black-box adversarial examples come from \texttt{Nat} and \texttt{Rand} sources.

When evaluating models against \texttt{White} adversarial examples at test-time, we use the test set which corresponds to their method of training. For instance, the \texttt{White} adversarial examples for the \texttt{Rand} model, come from the test set which has manipulated clean examples by \texttt{Rand} noise first. For \texttt{Vanilla}, \texttt{FIDS-W}, and \texttt{Ensemble} models, the adversarial examples are generated from clean data. This makes the comparison of models, which are trained on different types of data, fair. We use the one-shot attack for our white-box attack evaluation, using the same distribution based on natural noise.%, where all words are manipulated once.
 
  \subsection{Discussion}
Table \ref{robusttrans} shows the results for all models on all types of test data. Overall, our ensemble approach performs the best by a wide margin. As expected, adversarially-trained models usually perform best on the type of noise they have seen during training. However, we can notice that our \texttt{FIDS-W} model performs best on the \texttt{Nat} noise amongst models which have not been trained on this type of noise. Similarly, while  \texttt{FIDS-W} has not directly been trained on \texttt{Key} noise, it is trained on a more general type of noise, particularly flip, and thus can perform significantly better on the \texttt{Key} than on other models which also have not been trained on this type of noise. However, it cannot generalize to \texttt{Rand}, which is an extreme case of attack, and we need to use an ensemble approach to perform well on it too. Nevertheless, \texttt{FIDS-W} performs best on the \texttt{Rand} noise, compared with models which are not trained on \texttt{Rand} either. This validates our earlier claim that training on white-box adversarial examples, which are harder adversarial examples, can make the model more robust to weaker types of noise.
%Except for \texttt{Rand}, which randomly scramble all characters in a word, our one-shot \texttt{White} adversary brings the systems to the lowest BLEU score. In addition, compared with \texttt{Rand}, \texttt{White} performs many fewer manipulations. Interestingly,
 We also observe that \texttt{FIDS-B} performs better on the \texttt{White} examples compared with other baselines; although it has not been trained on white-box adversarial examples, it is trained on black-box adversarial examples of the same family of FIDS operations. 

\begin{figure}
\centering
\scalebox{0.4}{
\includegraphics[width=\linewidth]{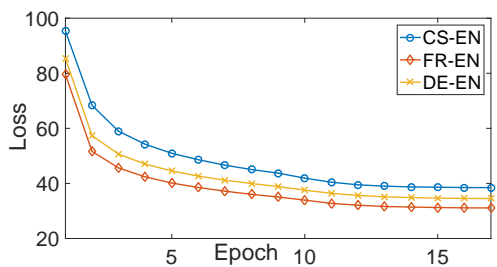}
}
\captionsetup{font=footnotesize}

\caption{Training loss on adversarial examples for \texttt{FIDS-W}.}\label{trainloss}
\end{figure}

Figure \ref{trainloss} shows the training loss of white-box adversarially-trained model on adversarial examples. The model is getting more resilient to adversarial examples, which are created at the start of each epoch.

  \begin{table*}[ht]
   \centering
 \scalebox{0.8}{
  \begin{tabular}{c|c|c|c|c|c|c|c|c }
   \multicolumn{2}{c|}{\backslashbox{Training}{Test}} & \texttt{Clean} & \texttt{Nat} & \texttt{Key}   &  \texttt{Rand} & \texttt{FIDS-B} & \texttt{FIDS-W} & Avg.  \\
    \hline
     \multirow{8}{*}{French} &  \texttt{Vanilla} &  37.54 &  19.17  &  12.12   & {4.75}    &  6.85  &  5.36  & 15.95          \\
      &  \texttt{Nat} & 26.35  &  \textbf{33.23}  &   11.16    &   5.32  &  8.28  &   6.65  & 15.16      \\
   
      % \hhline{~---------}
     &  \texttt{Key} & 33.02   &  17.30  &  \textbf{35.97}     &   4.63  & 7.00  &   5.17   & 17.17   \\
     %  \hhline{~---------}
      &  \texttt{Rand} & 36.06   & 18.54   &  8.31  &       \textbf{36.10}  & 8.76  &      7.14  & 19.14  \\
      % \hhline{~---------}
     &  \texttt{FIDS-B} &  34.48  &  21.59  &  28.48      &   6.82  &  32.62  &   \color{blue}{13.60 }  & 22.92  \\
      % \hhline{~---------}
    
      % \hhline{~---------}
     & \texttt{FIDS-W} &  37.15  &  \color{red}{23.65} &  \color{red}{31.18}   &       \color{red}{7.78}  &  \textbf{32.72}  &    \textbf{31.94}   &  27.40  \\    
      &  \texttt{Rand+Key+Nat} &  34.55 &  30.74 &   32.82   &   34.01  &  12.05  &    7.08  & 25.20  \\
       %\hhline{~---------}
     & \texttt{Ensemble}  & \textbf{37.81} &  30.27  &  29.36     &  34.42  &   32.00  &    {30.01}    & \textbf{32.30}  \\    
      \hline
      
      \multirow{8}{*}{German} &  \texttt{Vanilla} &  31.81 & 17.24   &   10.36 &     4.20    & 6.78   &   5.50   & 12.64      \\
       &  \texttt{Nat} &  24.89 &  \textbf{32.14} & 10.22   &  4.61 &  7.53       &     5.99  & 14.23  \\
      % \hhline{~---------}
     &  \texttt{Key} &  27.20  &   15.98  & \textbf{30.62}   & 4.64  &  7.68       &    4.74    & 15.13   \\
     %  \hhline{~---------}
        &  \texttt{Rand} &  31.01 &   17.90 &   6.59 &  \textbf{30.70}        &  9.19  &     5.83   & 16.86  \\

      % \hhline{~---------}
     &  \texttt{FIDS-B} & 28.27  & 20.22  &  23.84  & 6.29  &  27.35       &    \color{blue}{10.79}  & 19.45   \\
        & \texttt{FIDS-W} & \textbf{31.81}  & \color{red}{21.72}   &  \color{red}{26.23}   & \color{red}{7.75}  &   \textbf{27.38}  &       \textbf{26.51}  &  23.56      \\    
      % \hhline{~---------}
     &  \texttt{Rand+Key+Nat} & 29.22  &   29.78 &  27.83  &    28.88     &  10.30  & 6.14   &   22.01  \\
      % \hhline{~---------}
       %\hhline{~---------}
     & \texttt{Ensemble}  & 31.54   & 31.11   &  23.91  &   28.95 &  26.38       &   25.06  &  \textbf{27.82}    \\    
      \hline

\multirow{8}{*}{Czech} &  \texttt{Vanilla} &  \textbf{26.44}  &  13.55  &   9.49      &  4.78   & 7.30   &   5.93    & 11.24   \\
       &  \texttt{Nat} &  18.73 & \textbf{23.06}  &  9.07  &   4.45      &  7.36  &     5.42 & 11.34  \\
        &  \texttt{Key} & 22.76  & 13.09  &   \textbf{23.79} &  4.83       &  7.93  &   5.82 & 13.03    \\
     &  \texttt{Rand} & 24.23   &   12.00 &  7.26  & \textbf{24.53}  &    7.24     &   5.47   & 13.45  \\
      % \hhline{~---------}
    
     %  \hhline{~---------}
   
      % \hhline{~---------}
     &  \texttt{FIDS-B} & 22.31  &  14.15  &   17.91 &    6.48     &  19.67 &     \color{blue}{8.60}   & 14.84   \\
      & \texttt{FIDS-W} & 25.53  &   \color{red}{15.57} &    \color{red}{19.74}  &   \color{red}{7.18}       &  \textbf{20.02}  &    \textbf{19.42}  & 17.90  \\    
      % \hhline{~---------}
     &  \texttt{Rand+Key+Nat} & 22.21  &   20.59     &  20.60  &   21.33   &  10.06 &  5.89   & 16.77   \\
      % \hhline{~---------}
    
       %\hhline{~---------}
     & \texttt{Ensemble}  & 25.45  &  20.46 &  17.15       &  21.39    &  18.52  &   17.03   & \textbf{19.99}    \\    
      \hline

       \end{tabular}
    }
  \caption{BLEU score of models on clean and adversarial examples, using a decoder with beam size of 4. The best result on each test set is shown in bold. \texttt{FIDS-W} performs best on all noisy test sets compared with models which have not been trained on that particular noise (shown in red). \texttt{FIDS-B} performs best on white-box adversarial examples compared with other black-box trained models (shown in blue).} \label{robusttrans}
  \label{robustText}
\end{table*}

\section{Conclusion and Future Work}
%White-box attacks are among the most serious
%forms of attacks an adversary can inflict on a machine
%learning model. 
%We create white-box adversarial
%examples for a character-level NMT, and compare them with previously-studied black-box adversarial examples. We propose new types of attacks and introduce a new evaluation metric to compare adversaries with. We demonstrate the superiority of white-box adversarial examples to black-box adversarial examples in targeted and controlled scenarios. Due to efficiency of our adversarial example generation procedure, we are able to perform adversarial training, and thus create a more robust model, which performs better than black-box trained models on different types of noise.
%
%For future work, we will study other loss functions to explore our targeted adversary. For instance, we can incorporate attacks which have more than one word involved, and study cases where instead of masking non-target words, we can have a weighted objective function which includes more than one word.

As MT methods become more effective, more people trust and rely on their translations. This makes the remaining limitations of MT even more critical. Previous work showed that NMT performs poorly in the presence of random noise, and that its performance can be improved through adversarial training. We consider  stronger adversaries which are attacking a specific model and may also have specific goals, such as removing or changing words. Our white-box optimization, targeted attacks, and new evaluation methods are a step towards understanding and fixing the vulnerabilities in NMT: we're able to find more effective attacks and train more robust models than previous black-box methods. Next steps include exploring other types of targeted attacks, such as attacks that target more than one word, and other types of constraints, such as better characterizing which character changes affect intelligibility the least.

\section{Acknowledgment}
This work was funded by ARO grant W911NF-15-1-0265.

\bibliography{coling2018}
\bibliographystyle{acl}

\end{document}